\documentclass{article}

\PassOptionsToPackage{numbers, compress}{natbib}

\usepackage[nonatbib,final]{neurips_2019}

\usepackage[utf8]{inputenc} 
\usepackage[T1]{fontenc}    
\usepackage{hyperref}       
\usepackage{url}            
\usepackage{booktabs}       
\usepackage{amsfonts}       
\usepackage{nicefrac}       
\usepackage{microtype}      
\usepackage{amsmath}
\usepackage{amssymb}
\usepackage{graphicx}
\usepackage{graphics}
\usepackage{mathtools}
\usepackage{wrapfig}

\usepackage{dsfont}

\title{Hidden State Guidance: Improving Image Captioning Using an Image Conditioned Autoencoder}

\author{
   Jialin Wu \\
   Department of Computer Science\\
   University of Texas at Austin\\
  \texttt{jialinwu@utexas.edu} \\
   \And
   Raymond J. Mooney \\
   Department of Computer Science\\
   University of Texas at Austin\\
   \texttt{mooney@cs.utexas.edu} \\
}

\begin{document}

\maketitle

\begin{abstract}
Most RNN-based image captioning models receive supervision on the output words to mimic human captions. Therefore, the hidden states can only receive noisy gradient signals via layers of back-propagation through time, leading to less accurate generated captions. Consequently, we propose a novel framework, Hidden State Guidance (HSG), that matches the hidden states in the caption decoder to those in a teacher decoder trained on an easier task of autoencoding the captions conditioned on the image. During training with the REINFORCE algorithm, the conventional rewards are sentence-based evaluation metrics equally distributed to each generated word, no matter their relevance. HSG provides a word-level reward that helps the model learn better hidden representations. 
Experimental results demonstrate that HSG clearly outperforms various state-of-the-art caption decoders using either raw images or detected objects as inputs. 
\end{abstract}

Most captioning research \cite{xu2015show,donahue2015long,karpathy2015deep,vinyals2015show,anderson2017bottom,yao2018exploring,yang2018auto} trains an RNN-based decoder to learn the output word probabilities conditioned on the previous hidden state and various visual features. Recent methods improve results by incorporating richer visual inputs from object detection \cite{anderson2017bottom} and relationship detection \cite{yao2018exploring,yang2018auto}. 

By contrast, we focus on improving the hidden state representation learned during training. Most current image captioners are trained using maximum log-likelihood or REINFORCE with CIDEr \cite{vedantam2015cider} or BLEU \cite{Papineni:2002:BMA:1073083.1073135} rewards, where only the final word probabilities receive supervision. Therefore, the hidden states can only access noisy training signals from layers of backpropagation through time. Especially when training using REINFORCE, rewards are delayed until the end and equally distributed to each word in the caption, regardless of whether or not the words are descriptive, making the training signals even noisier. 
 
We present a new framework, called Hidden State Guidance (HSG), that treats the RNN caption decoder as a student network \cite{romero2014fitnets} and directly guides its hidden-state learning. However, this requires a teacher to provide hidden state supervision. We use a caption autoencoder as the teacher, giving it the same image as additional input. Its decoder has the same architecture as the caption decoder, allowing matching of the hidden states. Since the teacher has access to all of the human captions {\it and} visual inputs, its hidden states are expected to encode a richer representation that generates better captions, and therefore, provides useful hidden state supervision.
HSG plays a particularly helpful role when training using REINFORCE since it also provides a word-level intermediate reward that highlights the important words.
Our general framework can be used in almost any RNN-based image captioner. Experimental results show significant improvements over two recent caption decoders, FC \cite{rennie2017self}  using image features and Up-Down \cite{anderson2017bottom} using object-detection features.
\begin{figure*}[t]
\centering
\includegraphics[width=0.95\linewidth,trim={.8cm 8cm 4.8cm 0cm},clip]{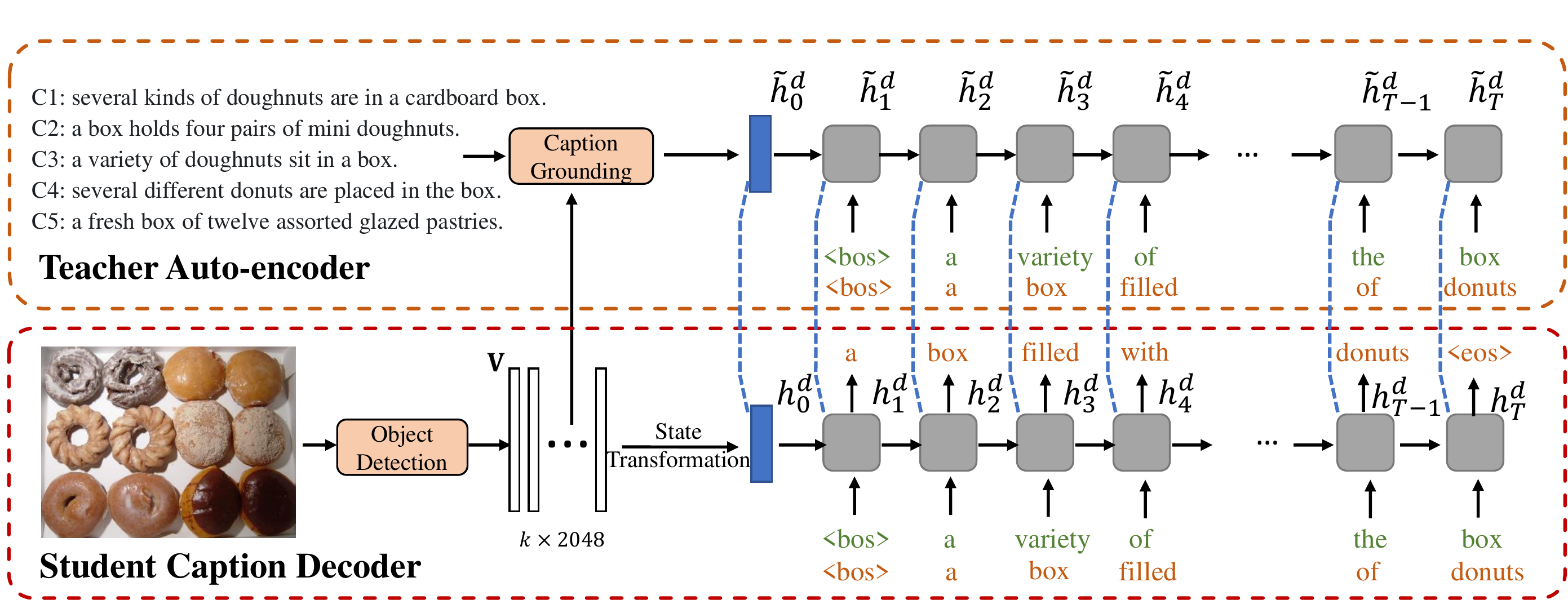}
\caption{Our framework consists of two parts. First, we train a teacher autoencoder that compresses the captions using the image as context. Second, the student decoder receives hidden state guidance from the teacher. Green captions present the maximum-likelihood training process, where each word from the human captions is fed to the network, and orange captions presents the REINFORCE case, where the previous generated word is fed to the caption decoder. Blue dashed lines indicate the hidden states' loss. }
\label{fig:model_overview}
\end{figure*}
\section{Approach}
We first present the overall architecture, then describe two student caption decoders, illustrating that HSG can be applied to almost any RNN-based decoder. After that, we explain the teacher autoencoder and the state transformation network that estimates the initial teacher hidden state from the visual inputs. 

\noindent\textbf{Overview}
\label{sec:overview}
The goal of HSG  is to provide hidden state guidance to any conventional RNN-based caption decoder, which we regard as a student network, as shown in Figure\ \ref{fig:model_overview}. In order to collect the guidance, we first train a teacher on an easier task that uses images to help autoencode human captions, which shares the same architecture as the student decoder. Then, we utilize a state transformation network to estimate the teacher decoder's initial hidden states ($t=0$) using only the visual input. These approximations are used to initialize the student decoder's hidden states so that it is capable of directly generating captions from images.

\noindent\textbf{Student Caption Decoder}
\label{sec:student}We briefly present two RNN-based student caption decoders.\\
\noindent\textbf{{FC}}. This model  \cite{vinyals2015show} adopts a single layer LSTM as the caption decoder. We first feed the full image to a deep CNN, and then average-pool the features from the final layer as visual features. The words are encoded using a trainable embedding matrix. At each time step, the LSTM receives the previous hidden states, generated words, and the visual features to generate the current word.\\
\noindent\textbf{Up-Down}. This model \cite{anderson2017bottom}  incorporates object detection features and has been widely adopted by recent research \cite{anderson2017bottom,wu2018faithful,yao2018exploring,yang2018auto}. The caption decoder operates on features of detected objects extracted using Faster RCNN \cite{girshick2015fast} with a ResNet-101 \cite{he2016deep} base network. It consists of a two-layer LSTM, where the first LSTM learns to distinguish important objects for generating the current word using an attention mechanism, and the second LSTM sequentially encodes the attended features to compute the output word probabilities.

\noindent\textbf{Teacher Autoencoder}.
Our teacher autoencoder is trained to generate captions using not only visual input features, but also the set of human captions for the image.

\noindent\textit{Teacher Caption Encoder}.
Our caption encoder takes as input the image feature set $\textbf{V}=\{\textbf{v}_1, ..., \textbf{v}_K\}$ consisting of $K$ vectors for $K$ detected objects, $C$ human captions $\textbf{W}^c_i = \{w^c_{i, 1}, w^c_{i, 2}, ..., w^c_{i,T}\}$, where $T$ denotes the length of the captions and $i=1,...,C$ are the caption indices.

\begin{wrapfigure}{l}{0.4\textwidth}
 \begin{center}
\includegraphics[width=0.4\textwidth,trim={0cm 8.5cm 22.5cm 0cm},clip]{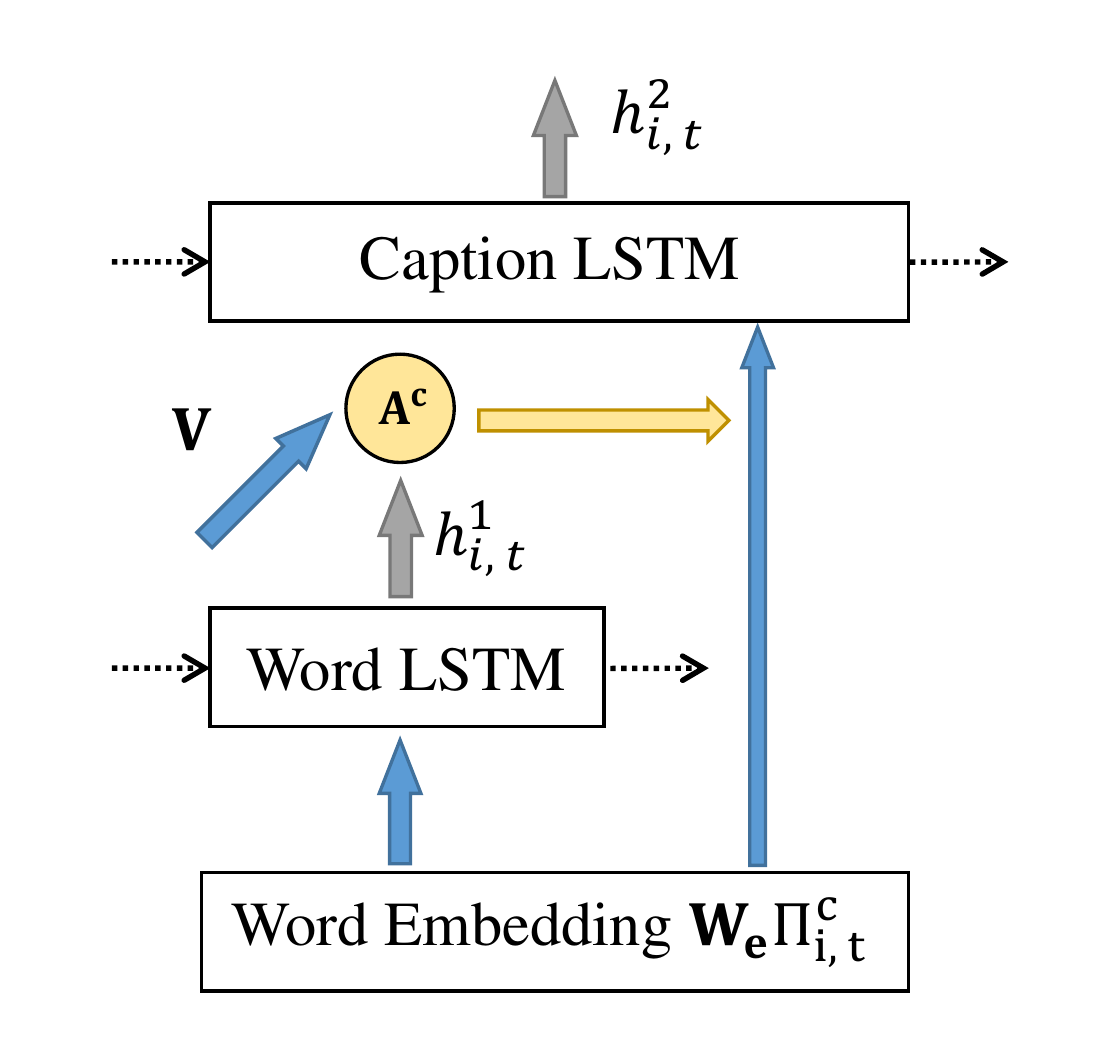}
 \end{center}
\caption{Overview of the caption encoder. The Word LSTM generates attention to identify the key words in each caption, and the Caption LSTM generates the final caption embedding. Blue arrows are fully-connected layers and yellow arrows are attention.}
\label{fig:caption_module}
\end{wrapfigure}

Inspired by \cite{wu2019generating}, we use a two-layer LSTM architecture  to encode human captions as illustrated in Figure\ \ref{fig:caption_module}. The first-layer LSTM (called the Word LSTM)  sequentially encodes the words in a caption $\textbf{W}^c_i$ at each time step as $h^{e,1}_{i,t}$: 
\begin{align}
    h^{e,1}_{i,t}, c^{e,1}_{i,t} &= \text{LSTM}( \textbf{W}_e \Pi^c_{i,t},~ h^{e,1}_{i, t-1}, ~c^{e,1}_{i,t-1}) 
    \label{eq:caption_word_attention}
\end{align}   
where $\textbf{W}_e$ is the 300-d word embedding matrix, and $\Pi^c_{i,t}$ is the one-hot embedding for the word $w^c_{i,t}$.

Then, we design a caption attention module $\textbf{A}^c$ which utilizes the image feature set $\textbf{V}^{q}$, and $h^{e,1}_{i,t}$ to generate the attention weight on the current word in order to indicate its importance. Specifically, the Word LSTM first encodes the word embedding $\Pi^c_{i,t}$ in Eq.\ \ref{eq:caption_word_attention}. Then we feed the outputs $h^{e,1}_{i,t}$ and $\textbf{V}$ to the attention module $\textbf{A}^c$. In particular, $\alpha^c_{i,j,t} = \text{softmax}( h^{e,1}_{i,t} \circ f(\textbf{v}_j)))$, where the softmax function is over the $K$ objects in visual feature set $\textbf{V}$.

Next, the attended word  representations $w_e$ in the caption are used to produce the final caption representation in Eq. \ref{eq:attended_caption_feed_foreward} via the Caption LSTM. 
\begin{align} 
&w_e =  \max_j \{\alpha^c_{i,j, t}\}\textbf{W}_e \Pi^c_{i,t}\\
\label{eq:attended_caption_feed_foreward}
    &h^{e,2}_{i,t}, c^{e,2}_{i,t} = \text{LSTM}(w_e, ~ h^{e,2}_{i,t-1}, ~ c^{e,2}_{i,t-1})
\end{align}
where $max$ denotes the element-wise max pooling over the attention weights for the $K$ objects.

\noindent\textit{Caption Decoder}. We require these two decoders to have the same architecture to provide hidden state guidance. The differences between these two decoders are the initial hidden states. The teacher decoder is initialized with the encoders' output while the student caption decoder is initialized with an estimated version.
For the FC  decoder, we use the max pooling of the final hidden state from the second LSTM in the caption encoder as the initial state. Similarly, we max pool the final hidden states from both layers to initialize the hidden states for the LSTMs in the Up-Down decoders.

\noindent\textbf{State Transformation Network}.
The state transformation network uses the visual features to estimate the initial teacher hidden states $ \tilde{h}^{d}_{0}$ so that the student caption decoder is capable of using the estimated hidden states to start a sentence purely from the visual inputs alone. For efficiency, we simply use a two-layer \textit{fc} network for state transformation. For the FC decoder, we directly apply the two-layer networks to the visual feature vector to estimate the initial hidden states  $ h^{d}_{0} = f(f(\textbf{v}))$.
\section{Training} 
We use $\theta_\alpha$ to denote the parameters in the autoencoder ($i.e.$ the caption grounding encoder and the teacher caption decoder), and  $\theta_g$ to denote the parameters in the state transformation network and the student decoder. We use $c$ to denote the entire caption, $c_t$ to denote the $t$-th word in the caption, and $c_{\leqslant t}$ to denote the first $t$ words in the caption. We omit the visual features $\textbf{v}$ in all of probabilities in this section for simplicity. We denote the maximum likelihood loss using parameters $\theta$ as $\mathcal{L}_{ll}(\theta)=-\sum^T_{t=1}\log(p(c_t|c_{\leqslant t-1};{\theta}))$.

\noindent\textbf{Pretraining the Teacher Autoencoder}.
We use cross-entropy loss, minimizing $\mathcal{L}_{ll}(\theta_\alpha)$. After pre-training, the parameters $\theta_\alpha$ are fixed. Additionally, we pre-train the state transformation network using $\mathcal{L}_{s,t}(\theta_g)= \|h^{d}_{t} - \tilde{h}^{d}_{t} \|^2_2$, $t$ = $0$. In particular, the generated captions from the student decoder are fed to the teacher autoencoder to compute the teacher hidden states at each time ($t$) as shown in Fig \ref{fig:model_overview}. We will omit ``$(\theta_g)$'' from $\mathcal{L}_{s,t}(\theta_g)$ for simplicity.

\noindent\textbf{Training the Student Decoder}.
We tested two different approaches to training the student decoders using either maximum likelihood or REINFORCE (with various evaluation metrics as rewards). The student decoder is initialized with the teacher decoder's parameters.

\noindent\textit{Maximum Likelihood Training}.
Maximum likelihood trains the student decoder to maximize the word-level log-likelihood, where human captions are fed into the decoder to compute the next word's probability distribution. We use the joint loss $\mathcal{L} =  \mathcal{L}_{ll}(\theta_g) + \lambda \sum_{t=0}^{T}\mathcal{L}_{s,t}$.
With human captions as input to the teacher autoencoder, we compute its hidden state, which is needed to calculate $\mathcal{L}_{s,t}$. The $\lambda$ parameter controls the weight of the state loss.

\noindent\textit{REINFORCE}. An alternative to log-likelihood maximization is to fine-tune the model to directly maximize the expected evaluation metric using REINFORCE. Negative rewards, such as BLEU \cite{Papineni:2002:BMA:1073083.1073135} or CIDEr \cite{vedantam2015cider}, are minimized using $\mathcal{L} =  - \mathbb{E}_{\hat{c} \sim p_{\theta_g}}[\tilde{r}(\hat{c})]$
where $\tilde{r}(\hat{c})=r(\hat{c}) - r(c^\star) $ denotes the variance-reduced rewards \cite{rennie2017self}, $\hat{c}$ denotes the sampled captions using the probabilities over the vocabulary, and  $c^\star$ denotes greedily sampled captions using the word with the maximum probability. We will omit ``$\theta_g$'' from $p_{\theta_g}$ for simplicity. The parameters in the student caption decoder are updated using the policy gradients $
\nabla_{\theta_g}\mathcal{L} = - \mathbb{E}_{\hat{c}\sim p} \Big[\tilde{r}(\hat{c}) \nabla_{\theta_g}\log p(\hat{c})\Big]\label{eq:policy_gradients}$.

\begin{table*}[t]
\centering
\small
\begin{tabular}{l|ccccc||ccccc}
\toprule
                              & \multicolumn{5}{c||}{Maximum Likelihood} & \multicolumn{5}{c}{REINFORCE\ (CIDEr)} \\ \hline
Model                     & B-4    & M    & R-L   & C     & S    & B-4    & M    & R-L    & C     & S   \\ \hline
LSTM-A   \cite{yao2017boosting}  &  35.2  & 26.9  & 55.8  & 108.8 &  20.0 &   35.5   & 27.3  & 56.8  & 118.3  & 20.8\\
StackCap \cite{gu2018stack}             & 35.2 & 26.5& - & 109.1& -  & 36.1& 27.4 & 56.9&  120.4 & 20.9  \\ \hline
FC \cite{vinyals2015show}      & 32.9  &  25.0  &  \textbf{54.0} & 95.4  & 17.9 & 32.8  & 25.0 & 54.2 & 104.0 & \textbf{18.5}\\
FC + HSG        & \textbf{33.2 } &  \textbf{25.5}  &   53.9 & \textbf{96.1} & \textbf{18.3} &\textbf{33.9}  & \textbf{25.9} & \textbf{54.8} & \textbf{107.5} & 18.4 \\\hline
Up-Down \cite{anderson2017bottom} &  \textbf{36.0} & 27.0 & 56.3 & 113.1 & 20.4 & 36.3 & 27.5 & 56.8 & 120.7 & 21.4 \\
Up-Down + HSG &  35.6 & \textbf{27.3} & \textbf{56.7} & \textbf{113.9} & \textbf{20.6} & \textbf{37.4} & \textbf{28.0} & \textbf{57.7} & \textbf{124.0} & \textbf{21.5} \\ \bottomrule

\end{tabular}
\caption{Automatic evaluation comparisons with various baseline caption decoders on the Karpathy test set.  ``HSG'' denotes trained with hidden state guidance, B-4, M, R-L, C and S are short hands for BLEU-4, METEOR, ROUGE-L, CIDEr and SPICE. All captions are generated with beam size 5.}
\label{tab:results}
\end{table*}

However, one remaining problem with this approach is that the sentence-level reward $\tilde{r}(\hat{c})$ is equally distributed over each word in the sampled captions, no matter how relevant the word is. Therefore, some desired words will not get enough credit because of the presence of some unrelated or inaccurate words in the sentence. To address this issue, we propose to use our hidden state loss as an intermediate reward to encourage the student decoder to produce  hidden states that match the hidden states of the high-performing teacher decoder. We add a reward objective $\tilde{\mathcal{R}} = -\sum\limits_{t=0}^{T}  \mathbb{E}_{\hat{c}_{ \leqslant t}\sim p}[\mathcal{L}_{s,t}]$ that is the accumulated expectation of the negative hidden state losses over time ($t$).
Therefore, the new policy gradients are: $\nabla_{\theta_g}\tilde{\mathcal{L}}= \mathbb{E}_{\hat{c}\sim p} [ \sum_{\tau=0}^{T}(\lambda \sum_{t=\tau}^{T} \mathcal{L}_{s,t}-\tilde{r}(\hat{c})) \nabla_{\theta_g} \log  p(\hat{c}_{\tau}| \hat{c}_{<\tau})] + \lambda \mathbb{E}_{\hat{c}\sim p}[ \sum_{t=0}^{T}  \nabla_{\theta_g}\mathcal{L}_{s,t} ]$.

It is worth noting that unlike the reward $\tilde{r}(\hat{c})$, the hidden state losses $\mathcal{L}_{s,t}$ are differentiable in the parameters $\theta_g$, which is necessary to compute the policy gradients. Intuitively, the new policy gradients can be understood as rewarding the student caption decoder when it produces hidden states that match the teacher's hidden states, and punishing it when the hidden states don't match. 

\section{Experimental Evaluation}
\noindent\textbf{Dataset}. 
We use the MSCOCO 2015 dataset \cite{chen2015microsoft} for image captioning. In particular, we use the Karpathy configuration that includes $110$K images for training and $5$K images each for validation and test. Each image has $5$ human caption annotations. We convert all sentences to lower case, tokenized on white spaces, and filter words that occur less than $5$ times. 

\noindent\textbf{Comparison with the Base Decoders}.
In Table \ref{tab:results}, we present the standard automatic evaluation for FC, Up-Down decoders trained using either Maximum Likelihood alone or using REINFORCE with CIDEr rewards. Metrics included are BLEU-4 \cite{Papineni:2002:BMA:1073083.1073135}, METEOR \cite{banerjee2005meteor}, ROUGE-L \cite{lin2004rouge}, CIDEr \cite{vedantam2015cider} and SPICE \cite{spice2016}. We observe a significant improvement on the CIDEr scores over all of the baseline models using REINFORCE (i.e. $107.5$ v.s. $104.0$ using FC Model, $124.0$ v.s. $120.7$ using FC Model). We attribute the improvements to both HSG and the word-level intermediate rewards. 
\section{Conclusion}
We have presented a  novel image captioning framework that uses an image-conditioned caption autoencoder. We observe that especially in the REINFORCE case, the word-level hidden state guidance assigns an intermediate reward that emphasizes the most relevant words. Extensive experimental results demonstrate the effectiveness of our approach. 
\section*{Acknowledgement}
This research was supported by the DARPA XAI program under a grant from AFRL. 
\bibliographystyle{ieee}
\bibliography{aaai}

\end{document}